# Incremental map generation by low cost robots based on possibility/necessity grids

Maite López-Sánchez     R. López de Mántaras     C. Sierra

IIIA-Artificial Intelligence Research Institute
CSIC-Spanish Scientific Research Council
Campus UAB, 08193 Bellaterra, Spain
mantaras@iiia.csic.es

## Abstract

In this paper we present some results obtained with a troupe of low-cost robots designed to cooperatively explore and adquire the map of unknown structured orthogonal environments. In order to improve the covering of the explored zone, the robots show different behaviours and cooperate by transferring each other the perceived environment when they meet. The returning robots deliver to a host computer their partial maps and the host incrementally generates the map of the environment by means of a possibility/ necessity grid.

## 1 INTRODUCTION

With the aim of exploring an structured environment that is unknown but easily passable, a troupe of low cost, small autonomous robots has been developed. These robots follow the already classical line of insect robots (Alami et al. 1993) (Brooks 1991). The goal of these autonomous robots is to explore and obtain partial information about an orthogonal environment and deliver this information to a host computer. Exploration is performed moving randomly and following walls (or obstacle edges) when detected. The computer host is expected to generate the most plausible map from the obtained information. This map models the environment in terms of degrees of possibility and necessity of the position of the detected walls and obstacles. The reason of choosing possibility/necessity techniques instead of probability is our need of an initial assigment of values representing ignorance. Possibility theory allows a clear representation of ignorance but probability does not. Regarding evidential theory, it is worth noticing that in our case Possibility and Necessity are in fact particular cases of Belief and Plausibility because our frame of discernment is $\Omega = \{wall, \neg wall\}$.

The behaviour of these small autonomous robots is similar -to some degree- to that of ants in two aspects. First, in order to increase the coverage of the environment, the robots have a partially random moving behaviour; and second, the robots cooperate by transferring each other the perceived environment when they meet. Sharing information in this way, allows the host to get the information not only from the robots that successfully return after an exploratory run, but also some information from those that could not return, provided that they had encountered robots that safely returned. Using this multi-robot strategy to generate a model of the environment, we expect to achieve a better efficiency than that which would be obtained based only on a single expensive robot.

The following section in this paper describes the structure and the behaviour of the robots. Then, we describe a statistical error analysis performed in order to know how the error intervals increase with the covered distance and the number of turns. This analysis will be used to model the environment by means of possibility/necesity techniques. The fourth section describes the map generation process based on the partial maps perceived by the successfully returning robots. Finally, we describe the results obtained to date, we briefly point to related work and we mention some future work.

## 2 STRUCTURE OF EACH MOVILE ROBOT

Each robot has been designed with the aim of being small and cheap. They must have a high autonomy and be endowed with a low cost processor to memorise the perceived environment map.

The robots environment perception system and the communication with the host or with other robots is based on IR impulse modulated sensors. The communication process consists of delivering the environmental information of a robot and it can be stablished between a robot and the host as well as between two robots that meet along their exploration. Therefore, this communication process allows to get all the information of non-returning mini robots that had been transferred to returning ones.

## 2.1 MECHANICAL CHARACTERISTICS

Each robot is 21 cm. length and 15 cm. wide (see Fig. 1). It has three wheels, two of them are 5 cm. steering wheels controlled by independent motors. The robots can reach a maximum speed up of 0.6 m/sec., and since the battery has about half hour of autonomy, each robot can do a maximum exploration of about 1000 m.

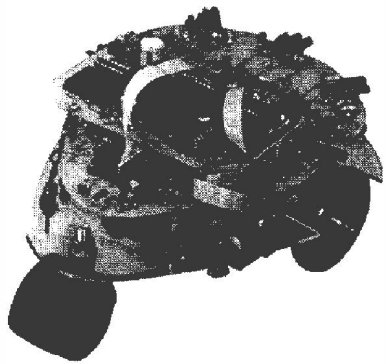

Figure 1: Autonomous Mini-Robot

## 2.1 SENSING CAPABILITY

Each robot is equipped with the following sensors:

- Impulse generators at each wheel for odometry.

- Five I.R. proximity sensors for frontal obstacles detection and for wall following.

- A proximity sensor for the detection of the terrain horizontal discontinuities.

- Safety micro switches for the detection of possible collision.

- One omnidirectional IR Emitter/Receiver sensor to detect other robots and to transmit data

- One IR Emitter with a scope of 90 degrees to generate a priority signal (right hand preference)

## 2.3 NAVIGATION STRATEGY

The navigation system incorporated to each robot has a partially random behaviour: The robot does a $\pm 45°$ or $\pm 90°$ turn either randomly or when it detects an obstacle.

The random turns are done with significantly different probabilities: $P_1 > P_2 > P_3$, corresponding to three differentiated behaviours:

$P_1$ = robot with an "Anxious" behaviour

$P_2$ = robot with "normal" behaviour

$P_3$ = robot with "routine" behaviour.

When the robot finds a frontal obstacle, the turn can be done to the right or to the left based also on a probability value $P_4$. The robots having a probability $P_4 < 0.5$ will show a tendency to turn to the right more often than to the left, whilst the robots having a probability $P_4 > 0.5$ will behave inversely.

Consequently, the different robots of the exploration troupe will not show an identical behaviour. They can behave in six different ways corresponding to the different combinations of behaviours and turning tendencies.

## 2.4 CONTROL SYSTEM

The control unit in each robot has been designed having in mind that the hardware had to be as simple as possible but, on the other hand, it had to allow achieving a behaviour sufficiently smart in order to navigate efficiently. Furthermore the robot had to be based on a hardware flexible enough to allow for experimentation of navigation and control strategies. These requirements have resulted in a design which contains three different functional modules : *the navigation module* that generates the trajectory to be followed; *the steering module* that controls the motors in order to follow the generated trajectory; and *the perception module* that acquires information of the environment by means of IR sensors. The computer used to implement the navigation control unit is a 80C186 with a 1MB RAM to store the perceived environment map. Finally, the steering control module is implemented on a 80C552 and operates with a resolution of 2 mm.

## 3 ERROR ANALYSIS

With the goal of studying the position error of each robot due to the imprecise odometry and to the imprecise steering, we have performed an analysis based on experimental data obtained from the real robots running straight (10 feet and 20 feet) and also turning 45 degrees left and 45 degrees right followed by a 10 feet straight run. We have performed 20 trials of each run and turning situation for each robot. With the data obtained, we have used the Kolmogorov normality test to verify that the experimental sample indeed follows a normal distribution both in the direction of the trajectory and in the direction perpendicular to the trajectory and we have tested that both distributions are independent. Based on this distributions we have determined the size of an error rectangle, comprising the 95% of the sample (which is elliptical shaped), associated to the final position of the robot after a straight run of 10 feet. This rectangle is 2.5 inches (in the direction of the trajectory) x 11 inches (in the direction perpendicular to the trajectory) in the average. We have also experimentally concluded that the size of the error rectangle is proportional to the covered distance. Concerning the additional error due to turning, we have obtained that when the robots turn 45 degrees there is, in the average, an error of about 2 degrees always towards the same direction. For example a robot with 2 degrees of error towards the left turns 43 degrees to the right instead of 45 degrees and turns about 47 degrees to the left instead of 45 degrees.

### 3.1 ERROR PROPAGATION

In free space, a trajectory is composed of a set of alternating segments and turns. Given the error rectangle at the initial point of a trajectory, we want to determine the error rectangle at the end of each segment taking into account the turning error and the error accumulated along the segment. The next figure shows the error propagation after a right turn, a straight line, another right turn and finally another straight line.

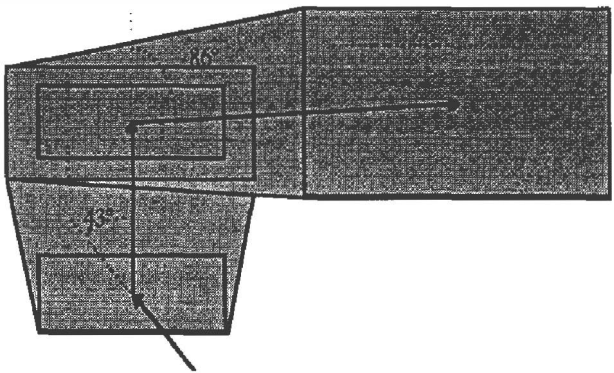

Figure 2: Error propagation

When following a wall, since the robot remains practically always at the same distance from the wall, the error along the direction orthogonal to the wall is taken to be constant and equal to the error that the robot has after turning to place itself parallel to the wall once the wall has been detected. This error analysis and error propagation study is performed on each robot and is used by the host to compute the possibility/necessity grid modelling the environment as described in the next section.

## 4 MAP GENERATION

The space being explored by the robots is discretized by means of a grid. Cells in the grid represent a small area of the real environment and contain two values : the degree of possibility and the degree of necessity of the presence of obstacles. Initially, that is before any exploration has taken place, all the cells have a possibility value $\Pi$ of 1 and a necessity value N of 0. These initial values correpond to a situation of total ignorance according to the theory of possibility (Dubois and Prade 1988). As robots communicate the information gathered during their exploration, the possibility and necessity values are modified in a way that depends on the presence, or not, of obstacles. The information gathered by each robot is nothing else but the trajectory of the robot together with the position of the walls detected (and followed) by the infrared sensors along the trajectory path, as well as the singular points detected, that is the wall ends and the corners. Due to the unavoidable odometry error, the position of the detected walls has an associated error. As we have explained in the last section, we have experimentally determined this error which has been approximated by a rectangle centered around the cell corresponding to the estimated position of the robot as shown in figure 3.

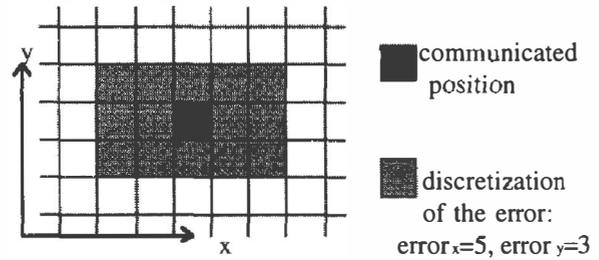

Figure 3: Grid representation of a position and its associated error.

Modelling the certainty of detected walls

When an error rectangle is associated to a position that belongs to a detected wall, the occupancy certainty degree (that is the certainty about the presence of an obstacle in that position) is expressed by means of necessity values in every cell that results partially or totally covered by the error rectangle around that position. The necessity values decrease linearly with the magnitude of the error and remains positive ($N(wall) = \alpha > 0$) in the cells inside the error rectangle but gets the value 0 at the cells outside the limits of the rectangle. These values have been established with the aim of reflecting that, having detected some obstacle, the necessity that there is a wall cannot be longer zero but positive since a positive value denotes some certainty degree about the occupancy of the space. However this occupancy certainty degree decreases when the distance to the central cell of the error rectangle increases. Figure 4 a) shows this case. Notice that the possibility value is constantly equal to 1 in all the cells covered by the error rectangle.

As we have already mentioned in the introduction, in our case Possibility and Necessity values are particular cases of Belief and Plausibility ones. We can easily see how our assigned values $N(wall) = \alpha > 0$ and $\Pi(wall) = 1$ can be considered as Belief(wall) and Plausibility(wall) corresponding to the following basic probability assignment (b.p.a.):
frame of discernment $\Omega = \{wall, \overline{wall}\}$,
with mass $m: P(\Omega) \rightarrow [0,1]$,
$\quad m(\emptyset)=0,\ m(wall)=\alpha,\ m(\overline{wall})=0,\ m(\Omega)=1-\alpha$.
and therefore, we obtain:

$$Bel(wall) = \sum_{A \subseteq wall} P(A) = m(wall) + m(\emptyset) = \alpha$$

$$Pl(wall) = \sum_{A \cap wall \neq 0} P(A) = m(wall) + m(\Omega) = 1 - \alpha$$

Modelling the certainty of free space

On the other hand, paths along which there was no detection supply information of free space, that is $\Pi(\neg wall)=1$ and $N(\neg wall)>0$, or equivalently, according to the axioms of possibility theory, $\Pi(wall) < 1$. This possibility value increases with the distance to the central cell of the error rectangle and reaches the value 1 at the cells outside the limits of the error rectangle. Obviously, we have $N(wall) = 0$ for all the cells covered by the error rectangle. Figure 4b shows this case.

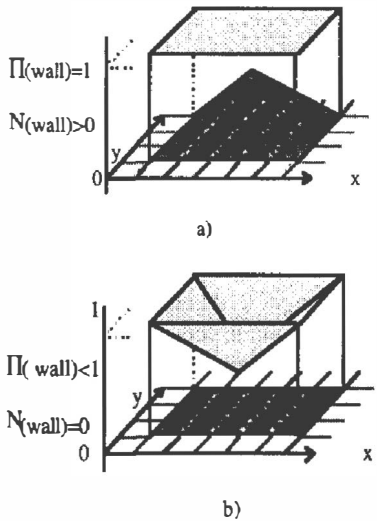

Figure 4: $\Pi$ and N values assigned to cells corresponding to: a) wall detection, and b) free space.

### 4.1 VALUE ASSIGNMENT

The height of the pyramids in figure 4 are determined by the magnitude of the error. The underlying idea is to establish a linear error-height relation such that, a null error implies the maximum allowed value of height (i.e. one), while an error too large implies a zero height since the information is no longer reliable. The error threshold that assigns a limit to a 'too large' error is established experimentally and is the same as the one that forces the robot to return from its exploration due to the irrelevancy of its later data. Summarising, the height values are obtained by applying the following formula:

$$height = 1 - \frac{current\ error}{maximum\ allowed\ error}$$

The computation of this height is done locally for each cell in the discretized environment grid on the basis of necessity propagation. Such propagation starts at the central cell and spreads over all those cells laying within the pyramid base. This is done passing four different values among cells: $e_r$, $e_l$, $e_u$ and $e_d$ which contain the distance between the current position and each side of the error rectangle, i.e. right, left, up and down respectively. This definition implies that their values are unitarily increased or decreased in each step of the propagation until they reach the zero value. Let $error_x$ be the length of the error rectangle base, and let $error_y$ be the rectangle height, then the error values are initially assigned at the central cell as follows: $e_r=e_l=error_x/2$ and $e_u=e_d=error_y/2$ and the following formulas are used to compute the height N corresponding to each cell within the error rectangle (Figure 5 shows schematically the propagation process): $N=min(N_x, N_y)$, where:

$$N_x = 1 - \frac{x}{err_x} - \frac{x - err_x}{max\ error}, \quad x = \frac{|e_l - e_r|}{2}, \quad err_x = \frac{e_l + e_r}{2}$$

$$N_y = 1 - \frac{y}{err_y} - \frac{y - err_y}{max\ error}, \quad y = \frac{|e_d - e_u|}{2}, \quad err_y = \frac{e_d + e_u}{2}$$

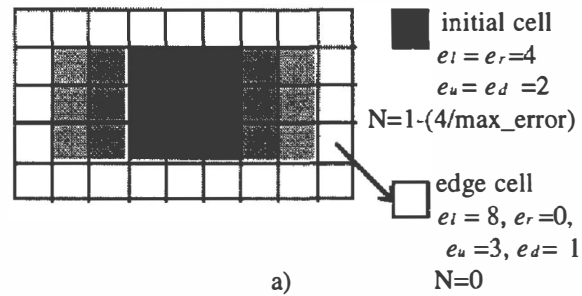

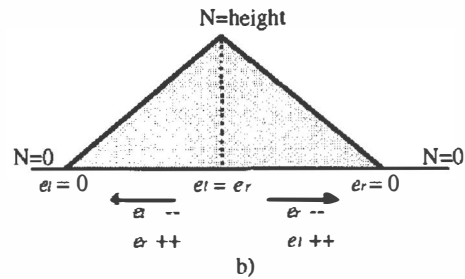

Figure 5: Value propagation: a) to adjacent cells, and b) along one dimension of the error rectangle.

### 4.2 COMBINATION OF VALUES

The cell necessity and possibility values representing trajectories in free space and wall segments are propagated from a central cell to the cells around as we have seen above. In considering consecutive points along the trajectory of the robot or along a wall segment, some of the cells covered by the current pyramid might already have values assigned by previous pyramids, and as a consequence the new values must be the result of a combination between these old and new values. In the case of wall segments the values are necessities (increasing from 0) and are combined by using the *max* operation (figure 6 b). In the case of trajectories these values are possibilities (decreasing from 1) and are combined by means of the *min* operator. Figure 6 a) graphically shows the results of such combination

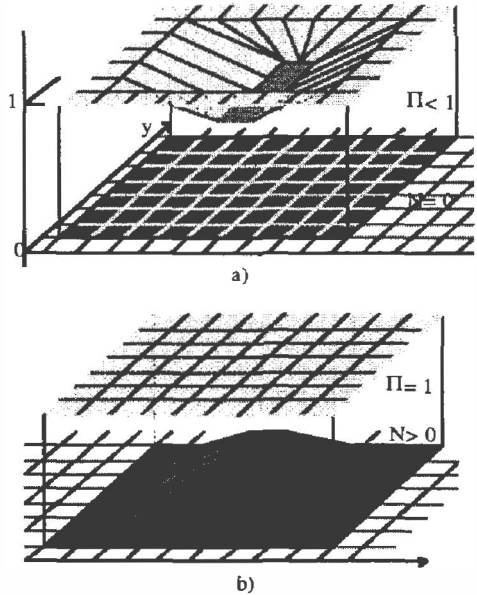

Figure 6: Segment representation corresponding to: a) trajectories (possibility pyramids), and b) walls (necessity pyramids).

When the same portion of a wall has been detected twice (or more) indepently, the necessities are combined by means of the probabilistic sum, that is $S(x,y) = x+y - xy$, in order to reinforce the certainty about the location of the wall. Figure 7 shows this situation

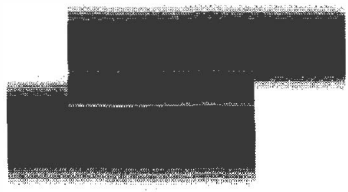

Figure 7: Reinforcement combination of two wall segments.

Following the interpretation of the Possibility/Necessity assignments as Belief/Plausibility values, we can justify now the use of the two different combination rules described above. On one hand, we have already seen that we apply the probabilistic sum when combining independent wall detections in the same cell, and this operation is nothing but Dempster's rule for simple support masses:

$Bel_i(wall) = \alpha_i \quad i = 1, 2$

$Bel_{1,2}^P(wall) = m_1 \oplus m_2(wall) = \alpha_1 + \alpha_2 - \alpha_1 \cdot \alpha_2$

$(and\ m(\Omega) = (1 - \alpha_1) \cdot (1 - \alpha_2))$

On the other hand, we also combine values coming from a single wall segment detection, and since we are considering non-independent evidences, Dempster's rule is not suitable for evidence combination. Instead, we have used a max-combination, a cautious operation whose results are still under the evidence theory framework. Indeed, max combination is in accordance with the so-called 'combination of compatible Belief functions' (Chateauneuf 1994) that makes sense when interpreting Bel/Pl values as bounds of the probability measures consistent with them. Namely, let

$F_{i=1,2} = \{P | Bel_i(A) \le P(A) \le Pl_i(A)\}$

be the family of such probabilities (Dempster 1967). Then, their natural combination can be taken as the intersection:

$F_{1,2} = F_1 \cap F_2 =$

$\{P | \max(Bel_1(A), Bel_2(A)) \le P(A) \le \min(Pl_1(A), Pl_2(A))\}$

In general, $\inf_{P \in F_1 \cap F_2} P(A)$ and $\sup_{P \in F_1 \cap F_2} P(A)$ are not a pair of Belief and Plausibility values (Chateauneuf 1994). However, in our particular case, this combination leads to a proper belief function. Indeed, the function $Bel^P$ is defined as

$Bel_{1,2}^P(wall) = \inf_{P \in F_1 \cap F_2} P(A) = \max(Bel_1(wall), Bel_2(wall)) =$
$= \max(\alpha_1, \alpha_2)$

$Bel_{1,2}^P(\emptyset) = Bel_{1,2}^P(\neg wall) = 0$

$Bel_{1,2}^P(\Omega) = 1$

a belief function whose corresponding mass assignments are:

$m(wall) = \max(\alpha_1, \alpha_2)$

$m(\emptyset) = m(\neg wall) = 0$

$m(\Omega) = 1 - \max(\alpha_1, \alpha_2)$

Moreover, in this particular case, this max-combination is also in accordance with a new combination operation proposed in (Torra 1995).

## 5 RESULTS

Figure 8 shows some of the results obtained, in simulation, with three robots departing from the point labelled "I" and taking into account the error position propagation along the trajectories. The orthogonal environment is represented by straight continuous lines, the trajectories by dark grey and the detected walls and obstacles by medium grey and the singular points by light grey. The darker the color along the three trajectories, the smaller the possibility value $\Pi$ of existence of a wall or obstacle. For the detected wall segments, the lighter the grey the smaller the certainty value N of the wall or obstacle being in that position. The grey degradation in the singular points also reflects the decrease of certainty about their actual position. The exploration stops when the cumulated error is higher than a previously set value.

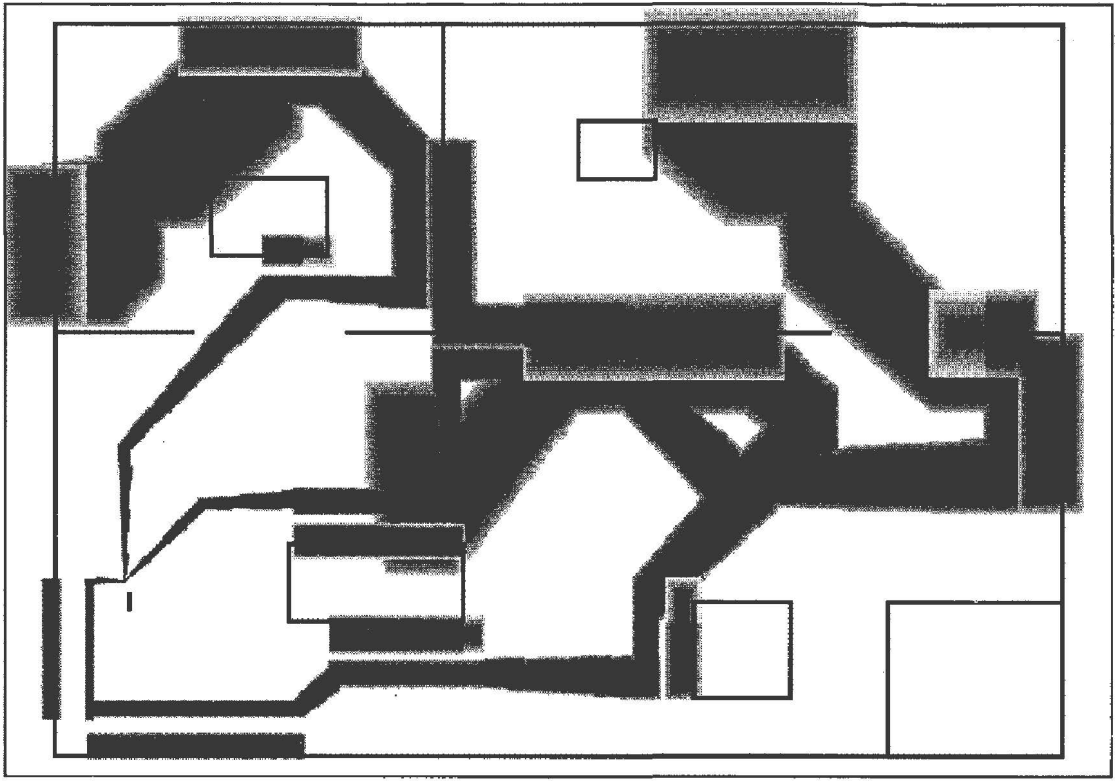

Figure 8: Global map obtained from three partial maps by three robots. 'I' indicates the exploration initial position of the robots.

## 6 RELATED WORK

There are quite a few works addressing the problem of map building. (Betge-Brezetz et al. 1996) use landmarks, defined as object features, to model natural environments and the uncertainty associated to their position is estimated by means of probabilistic techniques assuming a gaussian distribution of the uncertainty. In the case of certainty grid representations, the probabilistic approach has been also widely used to estimate the probability of cell occupancy (Moravec and Elfes 1985, Lim and Cho 1992, Pagac et al. 1996). Probabilistic techniques are reliable only if enough sensor data is available and, furthermore, if the data is well distributed in the explored environment and this distribution can be easily obtained. A very natural alternative when these conditions are not met is provided by fuzzy set theory. (Kim et al 1994) use fuzzy numbers to model the uncertainty of the parameters of geometric primitives and coordinate transformations used to describe natural environments. (Poloni et al. 1995) have also used fuzzy logic to build maps of unknown office-like environments. Their work is similar to ours in the sense that each point in the map has a degree of being empty and of being occupied however their approach uses straight fuzzy sets instead of dual possibility/necessity measures, another difference is that they work with only one robot and therefore no cooperation is involved, finally they use ultrasonic sensors instead of infrared ones and as a consequence the error accumulates faster than in our approach. The main consequence of working with only one robot and less precise sensors is that the maps built are significantly smaller.

## 7 CONCLUSIONS AND FURTHER WORK

The real robots are now working with a contour-based map building method also based on fuzzy techniques but we have detected some shortcomings due to the globality of the computational process involved, such shortcomings obliged us to adopt some ad hoc solutions during the proces of map completion (see Amat et al. 1995). The grid-based method presented here is completely based on a local computation process (the propagation of possibility and necessity values from a cell to their neighbours), exploits better the information about free space conveyed by the trajectories, takes advantage of the fact that possibility and necessity are dual measures and, furthermore, is computationally simpler. we are now in the process of incorporating this new approach to the real robots. On the other hand, further work is also in progress regarding the problem of planning additional trajectories towards zones of the environment poorly explored. In the long term we also plan to address the problem of learning higher level environment concepts ("corner", "door", etc.) based on sequences of sensor radings, i.e. we plan to address the problem of symbol grounding at least in simple orthogonal environments

## Acknowledgments

We would like to thank our colleague Dr. Lluis Godo for his helpful contributions and comments about the theoretical aspects of this work.

The robots have been designed and built under the supervision of Prof. Josep Amat at the LSI department (UPC, Barcelona, Spain).

## References.